\RequirePackage{fix-cm}
\documentclass[10pt,twocolumn,letterpaper]{article}
\usepackage[pagenumbers]{cvpr}
\usepackage[T1]{fontenc}
\usepackage{booktabs}
\usepackage{multirow}
\usepackage{placeins}

\usepackage{cuted} 

\usepackage{currfile} 

\usepackage{caption} 

\usepackage{bm}
\definecolor{cvprblue}{rgb}{0.21,0.49,0.74}
\usepackage[pagebackref,breaklinks,colorlinks,allcolors=cvprblue]{hyperref}
\def\confName{CVPR}
\def\confYear{2027}
\title{Semantic Guidance for Joint Infrared-Visible Video Fusion and Frame Interpolation}
\author{%
  Yushen Xu\textsuperscript{1}\quad
  Xiaosong Li\textsuperscript{2}\quad
  Huaming Wu\textsuperscript{1}\quad
  Dongwei Ren\textsuperscript{1}\thanks{Corresponding author.}\\
  \textsuperscript{1}Tianjin University\qquad
  \textsuperscript{2}Foshan University}

\begin{document}
\maketitle
\begin{strip}
    \centering
    \includegraphics[width=\linewidth]{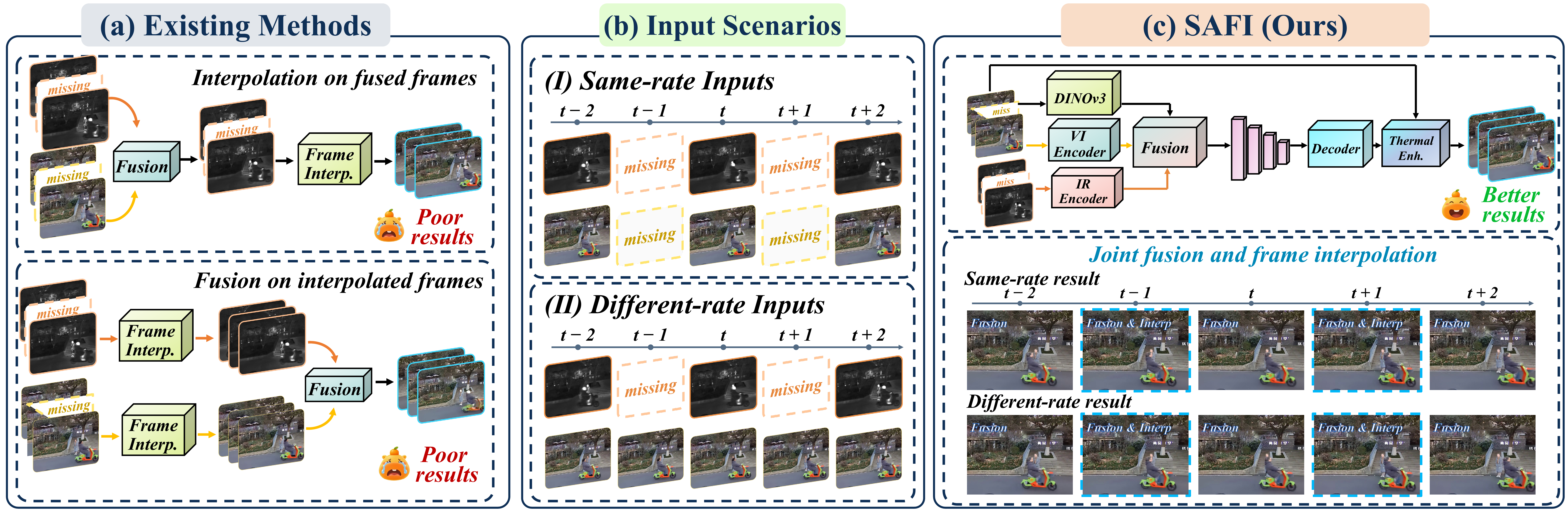}
    \captionsetup{hypcap=false}
    \captionof{figure}{\textbf{Motivation and paradigm comparison.}
(a) Existing methods that cascade fusion and interpolation in either order may accumulate errors and cause brightness fluctuations.
(b) Video fusion faces two missing-frame scenarios at target times requiring interpolation: both modalities lack frames with same-rate inputs, whereas only infrared frames are missing with different-rate inputs.
(c) For these two scenarios, SAFI jointly performs interpolation and fusion at the same target time, uses semantic guidance to preserve infrared target information and visible details, and constrains the temporal brightness consistency of its output.}
    \label{fig:fig1}
    \captionsetup{hypcap=true}
\end{strip}

\begin{abstract}
Infrared and visible video fusion at high frame rates requires recovering missing observations when camera rates differ or fall below the desired output rate. Cascading interpolation and fusion can accumulate errors and cause brightness fluctuations. Motion and occlusion further complicate content selection, particularly when missing visible frames also remove direct semantic guidance. We propose \textbf{SAFI}, which uses semantics at the target time to guide joint interpolation and fusion. When the visible frame rate exceeds the infrared rate, available visible frames provide semantic context directly. At times without visible observations, motion guides local matching, and propagation recovers semantics from neighboring frames. Semantic context complements aligned image features at the target time, while matching confidence helps select reliable infrared and visible information. Reconstruction and fusion share source selection within each scale, allowing supervision on missing frames to guide fusion. To preserve visible appearance and reduce flicker, we bound thermal enhancement and penalize brightness inconsistencies across frames while accounting for changes in thermal contrast. Experiments on three datasets show improved visual quality with less computation than cascaded baselines and consistent temporal gains when input rates differ. SAFI maintains strong downstream tracking performance despite missing frames.
The code is available at \url{https://github.com/XylonXu01/SAFI}.
\end{abstract}

\section{Introduction}
\label{sec:intro}
Nighttime driving benefits from fused videos that combine thermal targets with visible texture and color~\cite{guo2022antihalation,huang2026umvfuse}. In practical systems, unequal camera rates and limited temporal sampling leave one or both modalities without frames at the desired output times~\cite{ren2025meltpool}. High-frame-rate fusion must recover missing content and preserve complementary information despite motion and occlusion.

Fusion combines multimodal content~\cite{xu2026flexidfuse,xu2024tfsdiff}, whereas interpolation reconstructs missing observations. Semantic guidance provides scene context for choosing useful content~\cite{jie2025fsdiff}, but both tasks need this context at a common target time. Motion and occlusion make aligned observations unevenly reliable, and a missing visible frame also removes the direct source of semantic guidance.

SAFI obtains target-time semantics from an available visible frame or recovers them through motion-guided local matching and propagation. Motion narrows the search, while semantic similarity compares candidate content. Semantic features complement aligned image information, and matching confidence informs reliable source selection. Within each scale, shared aggregation weights let reconstruction supervision guide fusion's source selection.

To preserve visible appearance, SAFI converts the combined information into bounded thermal enhancement of a visible reference, preserving channel differences. Temporal supervision accounts for thermal changes while constraining low-frequency brightness fluctuations in motion-aligned errors. Our contributions are:
\begin{enumerate}
    \item We propose SAFI, which uses available or propagated target-time semantics to guide joint interpolation and fusion from incomplete infrared and visible observations in both frame-rate settings.
    \item Shared aggregation weights incorporate semantic matching confidence within each scale, allowing reconstruction supervision to guide fusion's source selection when aligned observations differ in reliability.
    \item Bounded thermal enhancement preserves visible channel differences, while temporal supervision accounts for changes in thermal contrast to reduce brightness flicker across consecutive fused frames.
\end{enumerate}

\section{Related work}
\label{sec:related_work}
\subsection{Infrared and visible video fusion}
Video fusion uses semantic, spatiotemporal, and historical cues for temporal consistency. TemCoCo~\cite{gong2025temcoco} distills DINOv2 semantics and VGG19 appearance, using inter-frame changes to preserve fused content and learn temporal variation. VideoFusion~\cite{tang2026videofusion} aggregates adjacent-frame information through bidirectional temporal attention and provides synchronized, spatially registered M3SVD for dynamic-scene fusion evaluation. Under extremely low illumination, USCFNet~\cite{zhao2026uscfnet} uses feature entropy to adapt spatiotemporal attention to visible reliability, guided by cross-modal content and edge differences. SDMFusion~\cite{liu2026sdmfusion} stores historical features in a single-step diffusion latent space, aligning and propagating them through temporal adapters. DRFusion~\cite{li2026drfusion} reduces autoregressive drift through history guidance and soft temporal anchoring. These methods seek consistent fusion across available input frames. With low frame rates or fast motion, large changes can occur between sampled frames, and fusion at observed times alone does not recover the missing intermediate content. Joint interpolation and fusion must reconstruct this content while assessing the differing reliability of aligned available and propagated target-time features.
\subsection{Video frame interpolation}
Frame interpolation combines motion estimation with aligned-frame aggregation. BiM-VFI~\cite{seo2025bimvfi} uses bidirectional fields for nonuniform motion, reducing intermediate-position ambiguity from temporal ratios alone. PerVFI~\cite{wu2024pervfi} uses sparse, nearly binary masks to blend primary and auxiliary frames asymmetrically, reducing double contours from alignment errors. Semantic and modality-specific cues support interpolation: ARVFI~\cite{ma2026arvfi} describes complex motion with DINOv3 semantics and autoregressively generates frames from both endpoints toward the middle. THERIS~\cite{zhou2026theris} interpolates infrared features with a heat-conduction prior, increasing temporal and spatial resolution. Joint interpolation and fusion must also determine how each modality's reconstructed features contribute to fusion.
\subsection{Joint frame interpolation and cross-modal fusion}
Image-level cascades use either processing order. Ren \etal~\cite{ren2025meltpool} transfer interpolation from high-frame-rate visible to low-frame-rate infrared videos, then fuse reconstructed infrared with corresponding visible frames. Guo \etal~\cite{guo2022antihalation} fuse selected frame pairs before timestamp-based, motion-compensated interpolation, reducing fusion operations and restoring original frame count and playback duration. Feature-level EVIF~\cite{geng2024evif} jointly optimizes event texture reconstruction, event-guided infrared deblurring, and fusion from visible events and infrared images. These image-level cascades can pass errors between stages, while EVIF requires visible events. SAFI uses target-time visible semantics to guide joint interpolation and fusion from two frame-based video streams, sharing image-source weights to select reliable evidence after motion compensation.

\section{Method}
\label{sec:method}

\begin{figure*}[t]
    \centering
    \includegraphics[width=\textwidth]{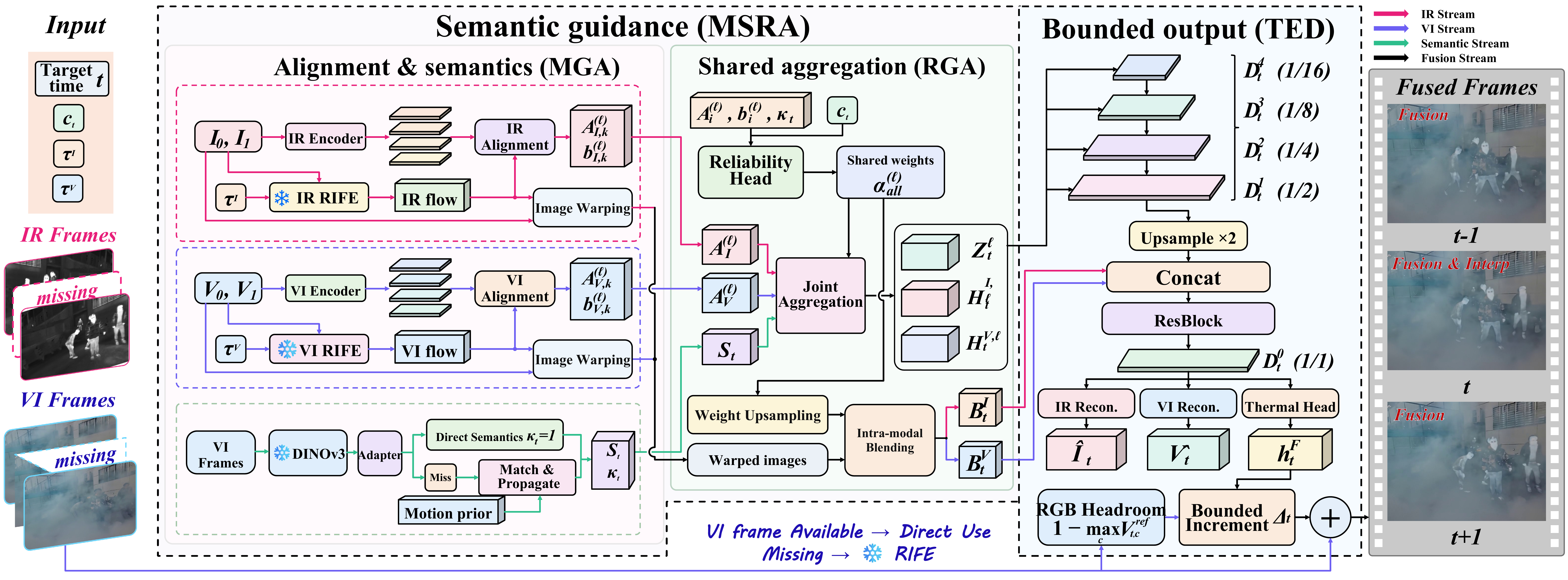}
    \caption{\textbf{Overview of SAFI.} Visible semantics guide joint interpolation and fusion at the target time. Semantic features complement aligned image features; matching confidence informs image weights shared within each scale, letting reconstruction supervision guide fusion. The combined features drive bounded thermal enhancement of a visible reference, preserving channel differences.}
    \label{fig:safi_framework}
\end{figure*}

SAFI combines target-time semantics with reliable image sources for joint interpolation and fusion, with constraints on thermal enhancement and temporal brightness (Fig.~\ref{fig:safi_framework}).

\subsection{Task definition and overview}
\label{subsec:problem}

We consider spatially registered infrared and visible sequences, $\mathcal{I}=\{(\mathbf{I}_i,t_i^I)\}$ and $\mathcal{V}=\{(\mathbf{V}_j,t_j^V)\}$, where $\mathbf{I}_i\in[0,1]^{H\times W\times1}$ and $\mathbf{V}_j\in[0,1]^{H\times W\times3}$.
We refer to acquired images supplied to the model as input frames, or as observed frames when emphasizing their acquisition status, to distinguish them from generated interpolated frames.
The target time $t$ is a time at which a fused frame is required, within the time range shared by both videos.
The target time can be an observed time, at which input frames from both modalities are available, or an interpolation time, at which at least one modality has a missing frame that requires reconstruction.
The network outputs the fused image $\mathbf{F}_t$ and reconstructs $\widehat{\mathbf{I}}_t$ and $\widehat{\mathbf{V}}_t$ through auxiliary branches.

Let $o_m(t)\in\{0,1\}$ indicate whether modality $m\in\{I,V\}$ has an input frame at the target time.
When a frame is available, the network uses it directly and masks the other input slots for that modality. Otherwise, it selects the input frames $\mathbf{X}_0^m,\mathbf{X}_1^m$ nearest to $t$ on either side.
Each target time uses at most four valid input sources, ordered as $\mathbf{I}_0,\mathbf{V}_0,\mathbf{I}_1,\mathbf{V}_1$.
The same-rate setting outputs fused frames at the infrared acquisition times and at the midpoints between consecutive acquisition times. When the visible frame rate exceeds the infrared frame rate, the network fills in infrared information along the visible timeline.
Both settings use the same network; only the input-frame times and validity states differ.

\subsection{Establishing target-time semantic guidance}
\label{subsec:msra}
\label{subsubsec:mga}

Neighboring frames can show the same object at different positions, whereas fusion requires information about the scene at a common target time. Motion-guided alignment (MGA) aligns image features and establishes semantic guidance at that time. It keeps the image sources separate so that their reliability can be assessed after alignment.

\paragraph{Aligning image sources.}
Modality-specific encoders extract multiscale image features. When modality $m$ has no frame at $t$, a frozen modality-specific RIFE branch~\cite{huang2022rife} estimates target-to-input sampling flows $\mathbf{u}_{t\rightarrow k}^{m}$ at the relative time $\tau_m=(t-t_0^m)/(t_1^m-t_0^m)$. Backward warping $\mathcal W$ aligns the features to $t$, yielding $\mathbf{A}_{m,k}^{\ell}$ with input and sampling validity $\mathbf{b}_{m,k}^{\ell}$. Available target-time inputs require no temporal warping.

\paragraph{Recovering target-time semantics.}
We obtain visible semantic features $\mathbf S_k$ by combining projected features from frozen DINOv3~\cite{simeoni2025dinov3}. An available target-time visible frame supplies semantics directly. Otherwise, semantics must be recovered from neighboring frames, whose content may be displaced. Motion estimates can be imperfect, so MGA uses motion to locate a candidate region and semantic similarity to compare the content within it. For a position $\mathbf p$ in the preceding frame, it matches valid candidates $\mathbf q$ in the following frame. Their set $\mathcal N_r$ is drawn from a local search window of half-width $r$ around the flow correspondence $\mathbf c_{\mathbf p}=\mathbf p+\mathbf u_{0\rightarrow1}^{V}(\mathbf p)$:
\begin{equation}
\begin{aligned}
C_{\mathbf p\mathbf q}
&=1-\cos(\mathbf S_0(\mathbf p),\mathbf S_1(\mathbf q))
+\eta d_{\mathbf p\mathbf q},\\
\pi_{\mathbf p\mathbf q}
&=\operatorname{softmax}_{\mathbf q\in\mathcal N_r}
(-C_{\mathbf p\mathbf q}/\epsilon).
\end{aligned}
\label{eq:semantic_matching}
\end{equation}
Here, $d_{\mathbf p\mathbf q}=\|\mathbf q-\mathbf c_{\mathbf p}\|_2^2/\max(1,r^2)$ penalizes deviations from the motion estimate; $\eta$ controls this penalty and $\epsilon$ is the matching temperature. Matching favors semantically similar content near the expected location, without changing the sampling flows.

The matched features update the preceding-frame semantics as $\mathbf S'_0(\mathbf p)=\tfrac12[\mathbf S_0(\mathbf p)+\sum_{\mathbf q}\pi_{\mathbf p\mathbf q}\mathbf S_1(\mathbf q)]$; the following-frame features $\mathbf S_1$ remain unchanged. Both are warped to $t$ and combined according to temporal position and sampling validity:
\begin{equation}
\begin{aligned}
\mathbf S_t&=\frac{(1-\tau_V)\mathbf b_0^S\odot\widetilde{\mathbf S}_{0,t}
+\tau_V\mathbf b_1^S\odot\widetilde{\mathbf S}_{1,t}}
{(1-\tau_V)\mathbf b_0^S+\tau_V\mathbf b_1^S+\varepsilon},\\
\widetilde{\mathbf S}_{0,t}&=\mathcal W(\mathbf S'_0,\mathbf u_{t\rightarrow0}^V),\\
\widetilde{\mathbf S}_{1,t}&=\mathcal W(\mathbf S_1,\mathbf u_{t\rightarrow1}^V).
\end{aligned}
\label{eq:semantic_propagation}
\end{equation}
The masks $\mathbf b_k^S$ retain valid samples. Matching confidence $\kappa_0(\mathbf p)=\max_{\mathbf q}\pi_{\mathbf p\mathbf q}$ is propagated with the preceding frame to $\bm\kappa_t$ for subsequent source selection; an available target-time visible frame has $\bm\kappa_t=1$. For propagated semantics, this confidence indicates how concentrated the local match is, providing a cue for judging the reliability of the aligned sources.

\subsection{Sharing reliable evidence across tasks}
\label{subsubsec:rga}
After alignment, motion errors and occlusion can still make some image sources less reliable than others. Fusion and missing-frame reconstruction both depend on selecting useful information from these sources. Reliability-guided aggregation (RGA) shares this selection through a common set of image-source weights, so reconstruction supervision also guides fusion's source selection. MGA and RGA form motion-guided semantic reliability aggregation (MSRA).

At each scale $\ell$, RGA uses aligned image features, input and sampling validity, semantic matching confidence $\bm\kappa_t$, and temporal conditions describing input timing and frame availability. It predicts raw visibility, reliability, and source scores $\mathbf v_i^\ell$, $\mathbf r_i^\ell$, and $\mathbf z_i^\ell$, together with a semantic-source score $\mathbf z_S^\ell$. Visibility and reliability adjust image-source contributions before joint normalization of the five source scores:
\begin{equation}
\begin{aligned}
\bm\nu_i^\ell&=\mathbf b_i^\ell\sigma(\mathbf v_i^\ell),\quad
\bm\rho_i^\ell=\bm\nu_i^\ell\sigma(\mathbf r_i^\ell),\\
\widetilde{\mathbf z}_i^\ell
&=\mathbf z_i^\ell+\log\max(\bm\rho_i^\ell,\varepsilon),\\
\bm\alpha_{\mathrm{all}}^\ell
&=\operatorname{softmax}\!\left(
[\widetilde{\mathbf z}_1^\ell,\ldots,\widetilde{\mathbf z}_4^\ell,\mathbf z_S^\ell]
\right).
\end{aligned}
\label{eq:reliability_weights}
\end{equation}
Here, $\sigma$ is sigmoid and $\varepsilon$ is a small constant for numerical stability. The score correction applies to valid sources; for invalid inputs or out-of-bounds samples ($\mathbf b_i^\ell=0$), $\widetilde{\mathbf z}_i^\ell=-\infty$, giving zero weight. Lower reliability penalizes image-source scores. The five entries of $\bm\alpha_{\mathrm{all}}^\ell$ are four image weights $\bm\alpha_i^\ell$ and one semantic weight $\bm\alpha_S^\ell$. Thus, matching confidence informs source selection, while target-time semantics supply complementary features for fusion.

The joint feature $\mathbf Z_t^\ell$ combines all five sources. For auxiliary reconstruction, $\mathbf H_t^{m,\ell}$ reuses the same image-source weights, renormalized within modality $m$:
\begin{equation}
\begin{aligned}
    \mathbf{Z}_t^\ell
    &=
    \sum_{i=1}^{4}\bm{\alpha}_i^\ell \mathbf{A}_i^\ell
    +\bm{\alpha}_S^\ell P_S^\ell(\mathbf{S}_t),\\
    \mathbf{H}_t^{m,\ell}
    &=
    \frac{
        \sum_{i\in\mathcal{S}_m}\bm{\alpha}_i^\ell \mathbf{A}_i^\ell
    }{
        \sum_{i\in\mathcal{S}_m}\bm{\alpha}_i^\ell+\varepsilon
    }.
\end{aligned}
\label{eq:joint_aggregation}
\end{equation}
Here, $P_S^\ell$ matches the semantic features to the current scale and channel dimension, and $\mathcal S_m$ indexes modality $m$'s inputs. Sharing keeps each modality's relative image-source weights consistent between the two aggregations. The weights are shared within each scale, while different scales use separate prediction heads.

\subsection{Bounding thermal enhancement}
\label{subsec:ted}
Fusion needs to retain visible appearance while adding infrared target information. We use a visible reference as the starting point and predict a bounded thermal increment from joint features. Auxiliary reconstruction trains the same features to recover missing-frame content.
\paragraph{Decoding joint features.}
The thermal enhancement decoder (TED) combines cross-modal features $\mathbf Z_t^\ell$, modality-specific reconstruction features $\mathbf H_t^{I,\ell},\mathbf H_t^{V,\ell}$, and pixel bases $\mathbf B_t^I,\mathbf B_t^V$ into full-resolution decoded features:
\begin{equation}
\mathbf D_t^0=\mathcal D\!\left(
\{\mathbf Z_t^\ell,\mathbf H_t^{I,\ell},\mathbf H_t^{V,\ell}\}_{\ell=1}^{4},
\mathbf B_t^I,\mathbf B_t^V
\right).
\label{eq:pyramid_decoder}
\end{equation}
As shown for TED in Fig.~\ref{fig:safi_framework}, the decoder $\mathcal D$ uses residual blocks from coarse to fine. Each finer-scale block concatenates the upsampled decoded features with the joint and reconstruction features at that scale. At full resolution, a final residual block combines the upsampled features with both pixel bases to produce $\mathbf D_t^0$. Three separate heads use these shared features to predict raw infrared and visible residuals $\mathbf h_t^I,\mathbf h_t^V$ and a single-channel thermal response $\mathbf h_t^F$.

\paragraph{Auxiliary reconstruction.}
The pixel base $\mathbf B_t^m$ is a modality-specific image estimate at $t$. It supplies pixel information to the decoder and provides a starting point for residual reconstruction. An available target-time frame provides the base directly; otherwise, motion-compensated inputs are combined using the shallowest-scale image-source weights, resized to image resolution and renormalized over valid sources within each modality. Missing frames use $\widehat{\mathbf X}_t^m=\operatorname{clip}(\mathbf B_t^m+0.2\tanh\mathbf h_t^m,0,1)$, while available inputs are returned as auxiliary outputs. Auxiliary reconstruction supervision trains the shared features and source weights to recover modality-specific content at missing target times.

\paragraph{Bounding output changes.}
The visible reference $\mathbf V_t^{\mathrm{ref}}$ is either the available target-time input or the refined RGB frame interpolated by frozen RIFE from neighboring inputs. It provides the RGB starting point for fusion; $\mathbf B_t^V$ supports decoding and auxiliary reconstruction. To limit the brightness change and keep RGB values within range, we bound the thermal increment by both its maximum amplitude and the reference's remaining headroom:
\begin{equation}
\begin{aligned}
    \bm{\Delta}_t
    &=
    \min\!\left(
        [0.25\tanh \mathbf{h}_t^F]_+,\,
        [1-\max_c\mathbf{V}_{t,c}^{\mathrm{ref}}]_+
    \right),\\
    \mathbf{F}_t
    &=\mathbf{V}_t^{\mathrm{ref}}+\bm{\Delta}_t\mathbf{1}_3.
\end{aligned}
\label{eq:fused_output}
\end{equation}
Here, $[x]_+=\max(x,0)$ and $\mathbf1_3$ broadcasts the increment to RGB. The nonnegative increment is bounded by both $0.25$ and the reference's headroom. Adding it equally to all channels preserves their differences and keeps the output within $[0,1]$. Joint features determine where and how much thermal enhancement is applied.

\subsection{Learning spatial and temporal consistency}
\label{subsec:losses}
Framewise reconstruction constrains content at each target time, but does not directly constrain changes between outputs. We therefore supervise spatial content, source selection, and motion-aligned temporal changes.
\paragraph{Reconstruction supervision.}
Fusion supervision ties thermal enhancement to the contrast between ground-truth infrared intensity and visible luminance. Given ground-truth frames $\mathbf I_t^*,\mathbf V_t^*$ and luminance $Y(\mathbf V)=0.299\mathbf V_R+0.587\mathbf V_G+0.114\mathbf V_B$, we define a bounded target $\bm\Delta_t^*$ and fused luminance error $\mathbf E_t=Y(\mathbf F_t)-Y(\mathbf V_t^*)-\bm\Delta_t^*$:
\begin{equation}
\begin{aligned}
\bm{\Delta}_t^*&=\min\!\Big(0.5[\mathbf{I}_t^*-Y(\mathbf{V}_t^*)]_+,\\
&\hspace{4em}0.25,\,1-\max_c \mathbf{V}_{t,c}^*\Big),\\
\mathcal{L}_{\mathrm{fus}}
&=2\|\bm{\Delta}_t-\bm{\Delta}_t^*\|_1+0.5\|\nabla \mathbf{E}_t\|_1.
\end{aligned}
\label{eq:fusion_loss}
\end{equation}
The two terms constrain enhancement strength and local luminance structure. To train the shared features to recover missing-frame content, we use
\begin{equation}
\mathcal L_{\mathrm{int}}=\mathcal L_I+3\mathcal L_V+0.1\mathcal L_{\nabla},
\label{eq:interpolation_loss}
\end{equation}
where $\mathcal L_I,\mathcal L_V$ are $L_1$ errors for missing infrared and visible frames, and $\mathcal L_{\nabla}$ averages their gradient errors.

\paragraph{Source supervision.}
For valid samples, a lower reconstruction error gives a higher reliability target. We set $\bm\rho_i^*=\mathbf b_i\exp(-\mathbf e_i/0.1)$ and $\bm\nu_i^*=\mathbf1[\bm\rho_i^*>0.2]$, where $\mathbf e_i$ is the motion-compensated error against ground truth and $\mathbf b_i$ indicates input and sampling validity. Binary cross-entropy (BCE) supervises the shallowest-scale reliability and visibility predictions:
\begin{equation}
\begin{aligned}
\mathcal L_{\mathrm{rel}}=\Big\langle
&\operatorname{BCE}(\uparrow\bm\rho_i^1,\bm\rho_i^*)\\
&+0.5\operatorname{BCE}(\uparrow\bm\nu_i^1,\bm\nu_i^*)
\Big\rangle_{\mathrm{act}}.
\end{aligned}
\label{eq:reliability_loss}
\end{equation}
Here, $\uparrow$ upsamples to image resolution, and $\langle\cdot\rangle_{\mathrm{act}}$ averages all pixels of valid input sources. Area averaging downsamples the reliability targets to $\bar{\bm\rho}_i^*$, which define the five source-weight targets:
\begin{equation}
\begin{aligned}
\bm{\alpha}_S^*
&=1-\max_i\bar{\bm{\rho}}_i^*,\\
\bm{\alpha}_i^*
&=(1-\bm{\alpha}_S^*)
\frac{\bar{\bm{\rho}}_i^*}
{\max(\sum_{j=1}^{4}\bar{\bm{\rho}}_j^*,\varepsilon)},\\
\mathcal L_{\mathrm{alloc}}
&=\left\langle
D_{\mathrm{KL}}(\bm\alpha^*\|\bm\alpha_{\mathrm{all}}^1)
\right\rangle_{\Omega_1}.
\end{aligned}
\label{eq:allocation_loss}
\end{equation}
As the most reliable image source weakens, the semantic target increases. The KL loss matches predicted weights to this target, with $\langle\cdot\rangle_{\Omega_1}$ averaging over shallowest-scale locations. To regularize semantic matching with the motion estimate, we define $q_{\mathbf p\mathbf q}=\operatorname{softmax}_{\mathbf q}(-d_{\mathbf p\mathbf q}/0.5)$ over valid local candidates $\mathcal N_{\mathbf p}^{\mathrm{val}}$ and use cross-entropy:
\begin{equation}
\mathcal L_{\mathrm{match}}
=-\frac{
\sum_{\mathbf p}M_{\mathbf p}
\sum_{\mathbf q\in\mathcal N_{\mathbf p}^{\mathrm{val}}}
q_{\mathbf p\mathbf q}\log\max(\pi_{\mathbf p\mathbf q},\varepsilon)
}{\max(\sum_{\mathbf p}M_{\mathbf p},\varepsilon)}.
\label{eq:matching_loss}
\end{equation}
Here, $M_{\mathbf p}$ weights valid supervision locations; the loss applies only when the target-time visible frame is missing. The targets and temporal teachers are used only for training; inference uses predicted source scores and weights.

\begin{figure*}[t]
    \centering
    \includegraphics[width=\textwidth]{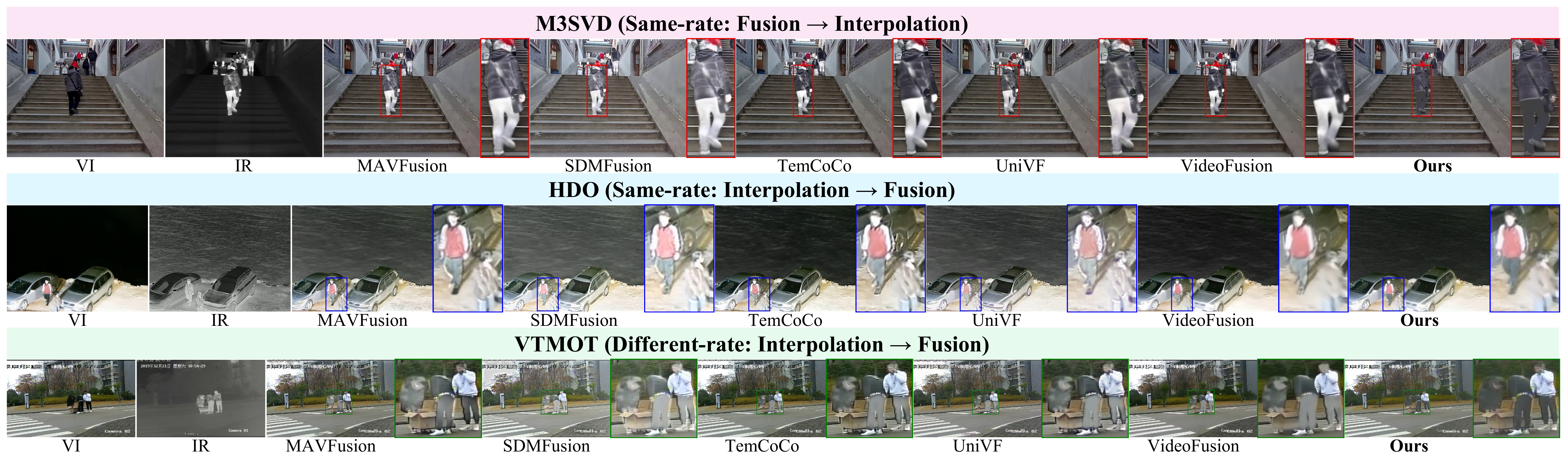}
    \caption{Fusion at interpolation times: same-rate fusion$\to$interpolation (top), same-rate interpolation$\to$fusion (middle), and different-rate interpolation$\to$fusion (bottom). SAFI performs both tasks jointly.}
    \label{fig:joint_qualitative}
\end{figure*}

\begin{table*}[t]
    \centering
    \definecolor{jointbest}{RGB}{237,43,43}
    \definecolor{jointsecond}{RGB}{25,65,180}
    \definecolor{jointthird}{RGB}{0,120,65}
    \newcommand{\jointbest}[1]{\textcolor{jointbest}{\textbf{#1}}}
    \newcommand{\jointsecond}[1]{\textcolor{jointsecond}{\underline{#1}}}
    \newcommand{\jointthird}[1]{\textcolor{jointthird}{#1}}
    \newcommand{\jointmetric}[2]{\mbox{\fontsize{7.5}{9}\selectfont#1$#2$}}
    \caption{Results at all observed and interpolation times. Baseline cascades: A, same-rate interpolation$\to$fusion; B, same-rate fusion$\to$interpolation; C, different-rate interpolation$\to$fusion. SAFI operates jointly. Best to third-best: \jointbest{bold red}, \jointsecond{underlined blue}, \jointthird{green}.}
    \label{tab:joint_results}
    \begingroup
    \footnotesize
    \setlength{\tabcolsep}{0.85pt}
    \renewcommand{\arraystretch}{1.0}
    \setlength{\aboverulesep}{1.2pt}
    \setlength{\belowrulesep}{1.2pt}
    \setlength{\cmidrulesep}{1pt}
    \begin{tabular*}{\textwidth}{@{\extracolsep{\fill}}cl*{6}{r}@{\hspace{5pt}}*{6}{r}@{\hspace{5pt}}*{6}{r}@{}}
        \toprule
        & & \multicolumn{6}{c}{HDO} & \multicolumn{6}{c}{M3SVD} & \multicolumn{6}{c}{VTMOT} \\
        \cmidrule(lr){3-8}\cmidrule(lr){9-14}\cmidrule(lr){15-20}
        Setup & Method & \jointmetric{SSIM}{\uparrow} & \jointmetric{$Q_{\mathrm{MI}}$}{\uparrow} & \jointmetric{VIF}{\uparrow} & \jointmetric{FSIM}{\uparrow} & \jointmetric{TCPE}{\downarrow} & \jointmetric{MMCI}{\downarrow} & \jointmetric{SSIM}{\uparrow} & \jointmetric{$Q_{\mathrm{MI}}$}{\uparrow} & \jointmetric{VIF}{\uparrow} & \jointmetric{FSIM}{\uparrow} & \jointmetric{TCPE}{\downarrow} & \jointmetric{MMCI}{\downarrow} & \jointmetric{SSIM}{\uparrow} & \jointmetric{$Q_{\mathrm{MI}}$}{\uparrow} & \jointmetric{VIF}{\uparrow} & \jointmetric{FSIM}{\uparrow} & \jointmetric{TCPE}{\downarrow} & \jointmetric{MMCI}{\downarrow} \\
        \cmidrule[\lightrulewidth](r){1-8}\cmidrule[\lightrulewidth](lr){9-14}\cmidrule[\lightrulewidth](l){15-20}
        \multirow{6}{*}{\textbf{A}} & MAVFusion & \jointthird{0.686} & \jointthird{0.430} & \jointsecond{0.260} & \jointsecond{0.788} & \jointthird{0.316} & 0.342 & \jointthird{0.577} & \jointsecond{0.544} & \jointbest{0.372} & \jointsecond{0.770} & \jointbest{0.244} & 0.175 & \jointthird{0.713} & \jointthird{0.532} & \jointsecond{0.362} & \jointthird{0.780} & \jointbest{0.302} & 0.287  \\
         & SDMFusion & 0.657 & 0.280 & \jointthird{0.220} & 0.775 & 0.321 & \jointsecond{0.185} & 0.536 & 0.395 & 0.291 & 0.751 & 0.264 & \jointsecond{0.128} & 0.695 & 0.387 & 0.296 & 0.774 & \jointsecond{0.314} & \jointbest{0.212}  \\
         & TemCoCo & 0.660 & 0.338 & 0.205 & 0.774 & 0.347 & 0.326 & 0.548 & 0.403 & 0.254 & 0.756 & 0.321 & 0.161 & 0.675 & 0.495 & 0.247 & 0.772 & \jointthird{0.372} & 0.275  \\
         & UniVF & \jointbest{0.692} & \jointsecond{0.445} & \jointsecond{0.260} & 0.785 & \jointsecond{0.313} & \jointthird{0.316} & \jointsecond{0.581} & \jointthird{0.520} & \jointthird{0.344} & \jointthird{0.768} & \jointthird{0.250} & \jointthird{0.139} & \jointsecond{0.720} & \jointsecond{0.578} & \jointthird{0.348} & 0.779 & \jointbest{0.302} & \jointthird{0.266}  \\
         & VideoFusion & 0.679 & 0.327 & \jointthird{0.220} & \jointthird{0.787} & 0.375 & 0.381 & \jointthird{0.577} & 0.437 & 0.253 & 0.767 & 0.342 & 0.196 & 0.709 & 0.523 & 0.279 & \jointsecond{0.786} & 0.397 & 0.296  \\
         & \textbf{SAFI} & \jointsecond{0.688} & \jointbest{0.646} & \jointbest{0.314} & \jointbest{0.792} & \jointbest{0.312} & \jointbest{0.131} & \jointbest{0.586} & \jointbest{0.707} & \jointsecond{0.358} & \jointbest{0.771} & \jointsecond{0.249} & \jointbest{0.115} & \jointbest{0.732} & \jointbest{0.723} & \jointbest{0.373} & \jointbest{0.795} & \jointbest{0.302} & \jointsecond{0.232}  \\
        \cmidrule[\lightrulewidth](r){1-8}\cmidrule[\lightrulewidth](lr){9-14}\cmidrule[\lightrulewidth](l){15-20}
        \multirow{6}{*}{\textbf{B}} & MAVFusion & \jointthird{0.687} & \jointthird{0.423} & \jointsecond{0.259} & \jointthird{0.787} & \jointthird{0.318} & 0.333 & 0.577 & \jointsecond{0.544} & \jointbest{0.373} & \jointsecond{0.770} & \jointbest{0.243} & 0.171 & \jointthird{0.714} & \jointthird{0.533} & \jointsecond{0.363} & \jointthird{0.780} & \jointbest{0.302} & 0.263  \\
         & SDMFusion & 0.657 & 0.280 & 0.221 & 0.775 & 0.324 & \jointsecond{0.183} & 0.536 & 0.395 & 0.291 & 0.751 & \jointthird{0.263} & \jointsecond{0.127} & 0.695 & 0.387 & 0.296 & 0.774 & \jointsecond{0.314} & \jointbest{0.202}  \\
         & TemCoCo & 0.661 & 0.333 & 0.200 & 0.776 & 0.349 & 0.318 & 0.549 & 0.404 & 0.256 & 0.757 & 0.313 & 0.156 & 0.677 & 0.496 & 0.250 & 0.773 & \jointthird{0.366} & 0.251  \\
         & UniVF & \jointbest{0.693} & \jointsecond{0.431} & \jointthird{0.258} & 0.785 & \jointsecond{0.314} & \jointthird{0.295} & \jointsecond{0.581} & \jointthird{0.521} & \jointthird{0.345} & \jointthird{0.768} & \jointsecond{0.249} & \jointthird{0.137} & \jointsecond{0.720} & \jointsecond{0.577} & \jointthird{0.348} & 0.779 & \jointbest{0.302} & \jointthird{0.243}  \\
         & VideoFusion & 0.682 & 0.329 & 0.220 & \jointsecond{0.789} & 0.370 & 0.377 & \jointthird{0.578} & 0.438 & 0.255 & 0.767 & 0.338 & 0.191 & 0.712 & 0.525 & 0.286 & \jointsecond{0.787} & 0.392 & 0.276  \\
         & \textbf{SAFI} & \jointsecond{0.688} & \jointbest{0.646} & \jointbest{0.314} & \jointbest{0.792} & \jointbest{0.312} & \jointbest{0.131} & \jointbest{0.586} & \jointbest{0.707} & \jointsecond{0.358} & \jointbest{0.771} & \jointsecond{0.249} & \jointbest{0.115} & \jointbest{0.732} & \jointbest{0.723} & \jointbest{0.373} & \jointbest{0.795} & \jointbest{0.302} & \jointsecond{0.232}  \\
        \cmidrule[\lightrulewidth](r){1-8}\cmidrule[\lightrulewidth](lr){9-14}\cmidrule[\lightrulewidth](l){15-20}
        \multirow{6}{*}{\textbf{C}} & MAVFusion & \jointthird{0.712} & \jointsecond{0.606} & \jointsecond{0.355} & \jointsecond{0.801} & \jointthird{0.193} & 0.088 & \jointthird{0.581} & \jointsecond{0.617} & \jointbest{0.441} & \jointsecond{0.772} & \jointsecond{0.192} & \jointsecond{0.063} & \jointthird{0.724} & \jointthird{0.607} & \jointsecond{0.445} & \jointthird{0.785} & \jointsecond{0.179} & \jointthird{0.117}  \\
         & SDMFusion & 0.683 & 0.327 & 0.290 & 0.787 & 0.205 & \jointthird{0.073} & 0.540 & 0.417 & 0.329 & 0.753 & 0.219 & 0.087 & 0.706 & 0.398 & 0.353 & 0.778 & \jointthird{0.200} & 0.146  \\
         & TemCoCo & 0.668 & 0.353 & 0.226 & 0.784 & 0.299 & 0.194 & 0.550 & 0.408 & 0.267 & 0.757 & 0.308 & 0.133 & 0.679 & 0.503 & 0.258 & 0.775 & 0.334 & 0.192  \\
         & UniVF & \jointbest{0.718} & \jointthird{0.583} & \jointthird{0.349} & \jointthird{0.800} & \jointsecond{0.189} & \jointsecond{0.047} & \jointsecond{0.585} & \jointthird{0.566} & \jointthird{0.397} & \jointthird{0.770} & \jointthird{0.204} & \jointthird{0.074} & \jointsecond{0.731} & \jointsecond{0.651} & \jointthird{0.427} & 0.784 & \jointsecond{0.179} & \jointsecond{0.101}  \\
         & VideoFusion & 0.686 & 0.339 & 0.233 & 0.793 & 0.347 & 0.344 & 0.579 & 0.441 & 0.261 & 0.768 & 0.336 & 0.190 & 0.713 & 0.530 & 0.294 & \jointsecond{0.788} & 0.376 & 0.263  \\
         & \textbf{SAFI} & \jointsecond{0.715} & \jointbest{0.959} & \jointbest{0.423} & \jointbest{0.809} & \jointbest{0.177} & \jointbest{0.013} & \jointbest{0.592} & \jointbest{0.898} & \jointsecond{0.415} & \jointbest{0.773} & \jointbest{0.191} & \jointbest{0.048} & \jointbest{0.743} & \jointbest{0.893} & \jointbest{0.448} & \jointbest{0.800} & \jointbest{0.175} & \jointbest{0.074}  \\
        \bottomrule
    \end{tabular*}
    \endgroup
\end{table*}

\begin{table*}[!t]
    \centering
    \caption{Efficiency in settings A--C (Table~\ref{tab:joint_results}). Baselines use GIMM-VFI-F. Parameters: total (fusion only in parentheses).}
    \label{tab:efficiency}
    \begingroup
    \fontsize{7}{8.5}\selectfont
    \setlength{\tabcolsep}{0.5pt}
    \renewcommand{\arraystretch}{1.05}
    \setlength{\aboverulesep}{1.2pt}
    \setlength{\belowrulesep}{1.2pt}
    \newcommand{\effparams}[2]{\mbox{#1 (#2)}}
    \begin{tabular*}{\textwidth}{@{\extracolsep{\fill}}lr*{3}{r}@{\hspace{5pt}}*{3}{r}@{\hspace{5pt}}*{3}{r}@{}}
        \toprule
        & & \multicolumn{3}{c}{\shortstack{A: Same-rate\\Interpolation$\to$Fusion}} & \multicolumn{3}{c}{\shortstack{B: Same-rate\\Fusion$\to$Interpolation}} & \multicolumn{3}{c}{\shortstack{C: Different-rate\\Interpolation$\to$Fusion}} \\
        \cmidrule(lr){3-5}\cmidrule(lr){6-8}\cmidrule(l){9-11}
        Method & \mbox{Params (M)$\downarrow$} & \mbox{FLOPs (G/frame)$\downarrow$} & \mbox{FPS$\uparrow$} & \mbox{Time (ms/frame)$\downarrow$} & \mbox{FLOPs (G/frame)$\downarrow$} & \mbox{FPS$\uparrow$} & \mbox{Time (ms/frame)$\downarrow$} & \mbox{FLOPs (G/frame)$\downarrow$} & \mbox{FPS$\uparrow$} & \mbox{Time (ms/frame)$\downarrow$} \\
        \midrule
        MAVFusion & \effparams{39.734}{9.149} & 7733.856 & 2.057 & 486.140 & 3872.046 & 4.072 & 245.590 & 3923.229 & 3.845 & 260.057 \\
        SDMFusion & \effparams{870.388}{839.803} & 9417.044 & 1.642 & 609.081 & 4732.780 & 3.233 & 309.319 & 5606.417 & 2.548 & 392.404 \\
        UniVF & \effparams{39.744}{9.158} & 9146.347 & 1.622 & 616.529 & 4642.496 & 3.175 & 314.949 & 5335.721 & 2.567 & 389.525 \\
        TemCoCo & \effparams{49.799}{19.213} & 9284.070 & 1.787 & 559.663 & 4642.035 & 3.555 & 281.306 & 5473.443 & 2.941 & 340.032 \\
        VideoFusion & \effparams{\textbf{37.336}}{\textbf{6.751}} & 8211.566 & 1.844 & 542.413 & 4164.814 & 3.560 & 280.879 & 4400.939 & 3.123 & 320.229 \\
        \midrule
        \textbf{SAFI} & 47.286 & \textbf{353.829} & \textbf{22.153} & \textbf{45.141} & \textbf{353.829} & \textbf{22.153} & \textbf{45.141} & \textbf{359.493} & \textbf{23.592} & \textbf{42.388} \\
        \bottomrule
    \end{tabular*}
    \endgroup
\end{table*}

\begin{figure}[t]
    \centering
    \includegraphics[width=\linewidth]{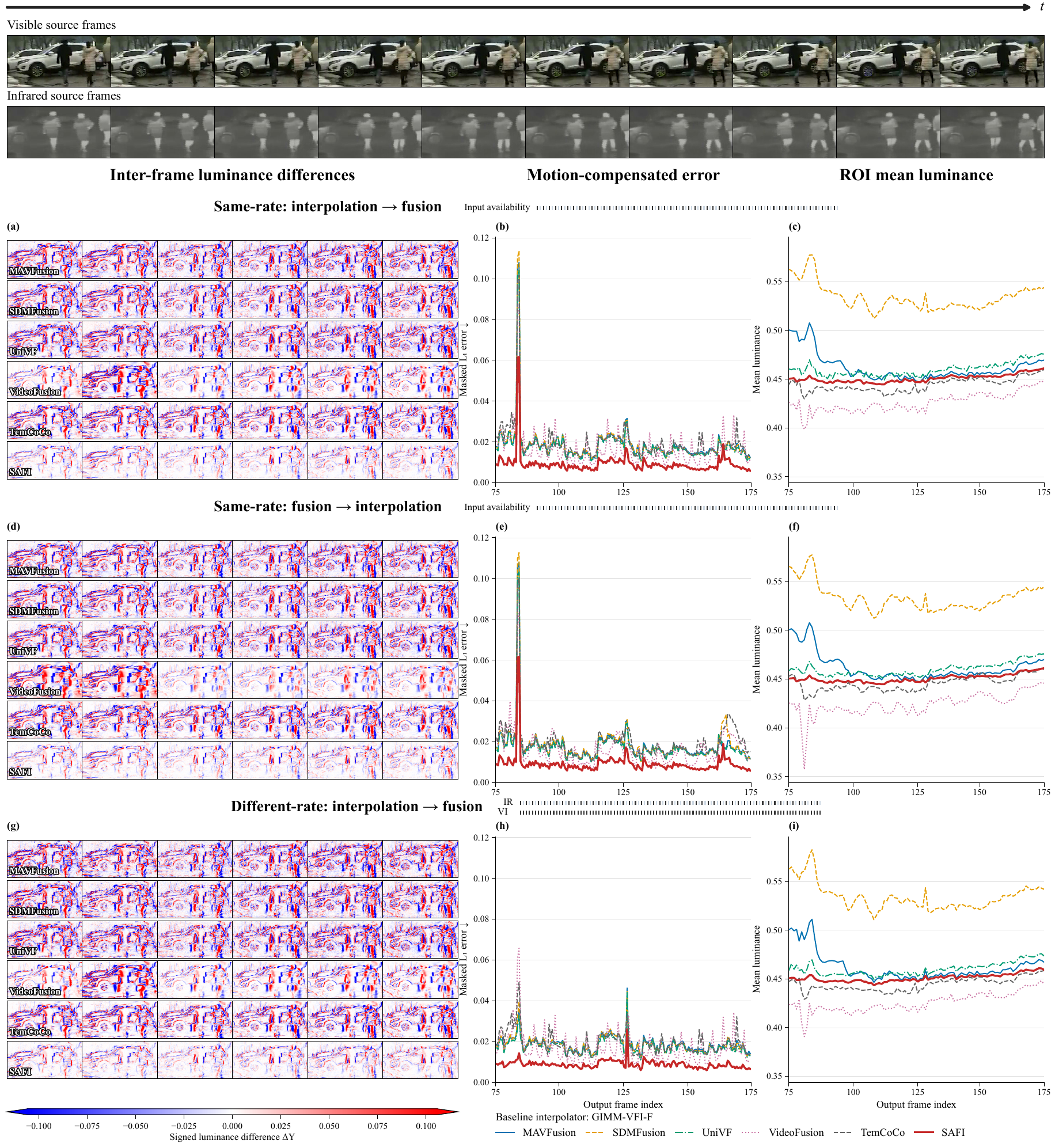}
    \caption{Temporal consistency in the three settings: (a, d, g) inter-frame luminance differences; (b, e, h) motion-compensated errors; (c, f, i) local mean luminance. Solid red curves: SAFI.}
    \label{fig:temporal_consistency}
\end{figure}

\begin{figure*}[t]
    \centering
    \includegraphics[width=\textwidth]{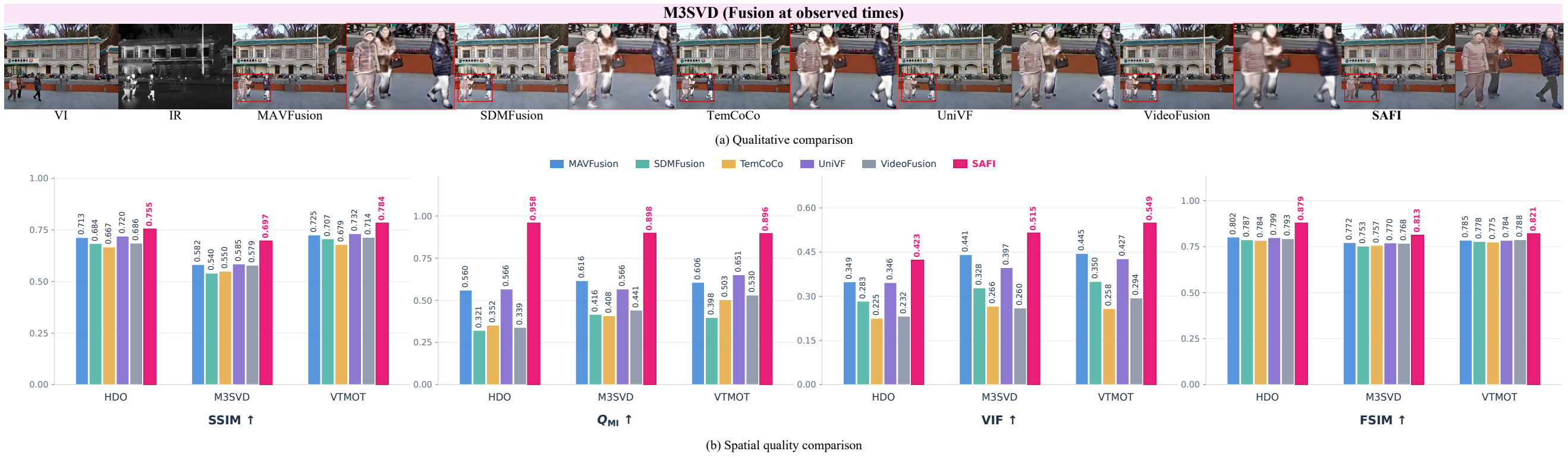}
    \caption{Fusion with observations from both modalities: (a) visual comparison; (b) spatial quality on HDO, M3SVD, and VTMOT.}
    \label{fig:observed_fusion}
\end{figure*}

\begin{table}[t]
    \centering
    \definecolor{trackbest}{RGB}{237,43,43}
    \definecolor{tracksecond}{RGB}{25,65,180}
    \definecolor{trackthird}{RGB}{0,120,65}
    \newcommand{\trackbest}[1]{\textcolor{trackbest}{\textbf{#1}}}
    \newcommand{\tracksecond}[1]{\textcolor{tracksecond}{\underline{#1}}}
    \newcommand{\trackthird}[1]{\textcolor{trackthird}{#1}}
    \caption{Tracking on VTMOT: baselines use complete videos; SAFI uses missing-frame inputs. Colors follow Table~\ref{tab:joint_results}.}
    \label{tab:tracking}
    \small
    \setlength{\tabcolsep}{2pt}
    \renewcommand{\arraystretch}{1.0}
    \setlength{\aboverulesep}{1.2pt}
    \setlength{\belowrulesep}{1.2pt}
    \begin{tabular*}{\linewidth}{@{\extracolsep{\fill}}l*{5}{r}@{}}
        \toprule
        Method & \mbox{HOTA$\uparrow$} & \mbox{MOTA$\uparrow$} & \mbox{IDF1$\uparrow$} & \mbox{DetA$\uparrow$} & \mbox{AssA$\uparrow$} \\
        \midrule
        Visible & 34.442 & 21.276 & 37.235 & 30.702 & \tracksecond{41.049} \\
        Infrared & 34.411 & 31.790 & 38.717 & 32.458 & 37.781 \\
        \midrule
        MAVFusion & \tracksecond{37.729} & 37.550 & \tracksecond{43.154} & \trackthird{37.488} & 39.310 \\
        SDMFusion & \trackthird{37.726} & \tracksecond{37.813} & \trackthird{42.930} & \tracksecond{37.599} & 39.182 \\
        UniVF & 37.201 & \trackthird{37.599} & 42.380 & 36.492 & \trackthird{39.390} \\
        TemCoCo & 34.949 & 32.654 & 39.047 & 34.612 & 36.691 \\
        VideoFusion & 34.052 & 31.871 & 38.290 & 32.867 & 36.713 \\
        \midrule
        \textbf{SAFI} & \trackbest{43.085} & \trackbest{43.790} & \trackbest{48.874} & \trackbest{41.451} & \trackbest{46.825} \\
        \bottomrule
    \end{tabular*}
\end{table}

\begin{table*}[t]
    \centering
    \caption{Aggregated ablation results on HDO, M3SVD, and VTMOT. Bold: best within each frame-rate setting.}
    \label{tab:ablation}
    \small
    \setlength{\tabcolsep}{2pt}
    \renewcommand{\arraystretch}{1.0}
    \setlength{\aboverulesep}{1.2pt}
    \setlength{\belowrulesep}{1.2pt}
    \begin{tabular*}{\textwidth}{@{\extracolsep{\fill}}l*{6}{r}@{\hspace{12pt}}*{6}{r}@{}}
        \toprule
        & \multicolumn{6}{c}{(a) Same-rate setting} & \multicolumn{6}{c}{(b) Different-rate setting} \\
        \cmidrule(lr){2-7}\cmidrule(lr){8-13}
        Model & \mbox{SSIM$\uparrow$} & \mbox{$Q_{\mathrm{MI}}\uparrow$} & \mbox{VIF$\uparrow$} & \mbox{FSIM$\uparrow$} & \mbox{TCPE$\downarrow$} & \mbox{MMCI$\downarrow$} & \mbox{SSIM$\uparrow$} & \mbox{$Q_{\mathrm{MI}}\uparrow$} & \mbox{VIF$\uparrow$} & \mbox{FSIM$\uparrow$} & \mbox{TCPE$\downarrow$} & \mbox{MMCI$\downarrow$} \\
        \midrule
        \textbf{Full model} & \textbf{0.700} & \textbf{0.714} & \textbf{0.362} & \textbf{0.790} & \textbf{0.297} & \textbf{0.159} & \textbf{0.712} & \textbf{0.893} & \textbf{0.439} & \textbf{0.796} & \textbf{0.178} & \textbf{0.068} \\
        A1 & 0.526 & 0.471 & 0.272 & 0.437 & 0.372 & 0.372 & 0.370 & 0.555 & 0.296 & 0.542 & 0.267 & 0.090 \\
        A2 & 0.511 & 0.403 & 0.287 & 0.426 & 0.389 & 0.364 & 0.552 & 0.705 & 0.226 & 0.506 & 0.264 & 0.096 \\
        A3 & 0.495 & 0.514 & 0.255 & 0.436 & 0.384 & 0.378 & 0.424 & 0.626 & 0.315 & 0.498 & 0.244 & 0.122 \\
        A4 & 0.480 & 0.514 & 0.181 & 0.436 & 0.373 & 0.307 & 0.520 & 0.681 & 0.239 & 0.532 & 0.240 & 0.087 \\
        \bottomrule
    \end{tabular*}
\end{table*}

\paragraph{Temporal supervision.}
Forcing a constant enhancement increment would suppress real changes in thermal contrast. Temporal consistency loss (TCL) accounts for changes in the thermal increment target derived from ground-truth frames when supervising adjacent outputs. For an anchor time $s$ with an available visible frame, frozen infrared teacher flow aligns the predicted increment, which is corrected by the change in this target:
\begin{equation}
\begin{aligned}
\widetilde{\bm{\Delta}}_{s\rightarrow t}=\Pi_t\Big[
&\mathcal W^I_{s\rightarrow t}(\operatorname{sg}(\bm{\Delta}_s))\\
&+\bm{\Delta}_t^*-\mathcal W^I_{s\rightarrow t}(\bm{\Delta}_s^*)\Big].
\end{aligned}
\label{eq:temporal_target}
\end{equation}
Here, $\operatorname{sg}$ stops gradients at the anchor, and $\Pi_t$ clips to $[0,\min(0.25,1-\max_c\mathbf V_{t,c}^*)]$. The enhancement error is $\mathbf R_{st}^I=\bm\Delta_t-\widetilde{\bm\Delta}_{s\rightarrow t}$. To address broader brightness fluctuations, we use frozen visible teacher flow to compute the aligned luminance-error difference $\mathbf R_{st}^V=\mathbf E_t-\mathcal W^V_{s\rightarrow t}(\mathbf E_s)$ and penalize its low-frequency component after weighting:
\begin{equation}
\begin{aligned}
\mathcal L_{\mathrm{temp}}&=
\operatorname{Avg}_{\mathcal P_\Delta}
\frac{\sum_{\mathbf p}\mathbf{G}_{st}^I\mathbf{C}_{st}^I|\mathbf{R}_{st}^I|}
{\sum_{\mathbf p}\mathbf{G}_{st}^I+\varepsilon},\\
\mathcal L_{\mathrm{lf}}&=
\operatorname{Avg}_{\mathcal P_{\leftrightarrow}}
\frac{\|\mathcal A(\mathbf{G}_{st}^V\mathbf{C}_{st}^V\odot \mathbf{R}_{st}^V)\|_1}
{\|\mathcal A(\mathbf{G}_{st}^V)\|_1+\varepsilon}.
\end{aligned}
\label{eq:low_frequency_loss}
\end{equation}
Validity $\mathbf G_{st}^m$ and confidence $\mathbf C_{st}^m$ weight the aligned errors, and $\mathcal A$ is $8\times8$ average pooling. The averages use valid adjacent-time pairs: $\mathcal P_\Delta$ requires an available visible anchor, whereas $\mathcal P_{\leftrightarrow}$ uses both directions with gradients at both endpoints.

The training objective is
\begin{equation}
\begin{aligned}
\mathcal L={}&\mathcal L_{\mathrm{int}}+\mathcal L_{\mathrm{fus}}
+0.1\mathcal L_{\mathrm{rel}}\\
&+0.05\mathcal L_{\mathrm{alloc}}+0.05\mathcal L_{\mathrm{match}}\\
&+0.1g(k)(\mathcal L_{\mathrm{temp}}+\mathcal L_{\mathrm{lf}}),
\end{aligned}
\label{eq:total_loss}
\end{equation}
where $g(k)=0.1+0.9\min(k/500,1)$ increases the temporal weight and $k$ counts completed updates.

\section{Experiments}
\label{sec:experiments}

\subsection{Experimental setup}
\label{sec:experimental_setup}

\noindent\textbf{Datasets and protocols.}
We use HDO~\cite{xie2024rcvs}, M3SVD~\cite{tang2026videofusion}, and VTMOT~\cite{zhu2025vtmot}, with 589/66/157 training/validation/test sequences. Same-rate inputs retain every other frame of both modalities; the fused output doubles their frame rate. Different-rate inputs retain every other infrared frame and all visible frames; the output matches the visible frame rate, twice the infrared rate.

\noindent\textbf{Comparison methods.}
We compare with MAVFusion (ECCV'26)~\cite{li2026mavfusion}, SDMFusion (CVPR'26)~\cite{liu2026sdmfusion}, UniVF (NeurIPS'25)~\cite{zhao2025univf}, VideoFusion (CVPR'26)~\cite{tang2026videofusion}, and TemCoCo (ICCV'25)~\cite{gong2025temcoco}, each cascaded with GIMM-VFI-F (NeurIPS'24)~\cite{guo2024generalizable}. We choose GIMM-VFI-F over RIFE for better interpolation quality. We test both cascade orders for same-rate inputs and interpolation followed by fusion for different-rate inputs, where missing paired observations prevent fusion as the first step. Due to limited space, interpolation-only comparisons appear in the appendix.

\noindent\textbf{Evaluation metrics.}
Using the original target-time infrared and visible images as references, we measure spatial quality with SSIM~\cite{wang2004ssim}, $Q_{\mathrm{MI}}$~\cite{hossny2008comments}, VIF~\cite{sheikh2006vif}, and FSIM~\cite{zhang2011fsim}, and temporal consistency with TCPE and MMCI~\cite{li2026ftpfusion}.

\noindent\textbf{Implementation details.}
Training uses both frame-rate protocols and shared parameters on three NVIDIA RTX 4090 GPUs. Official pretrained DINOv3 ViT-S/16~\cite{simeoni2025dinov3} provides semantics. We use AdamW~\cite{loshchilov2019adamw}, $256\times256$ crops, and an initial learning rate of $10^{-5}$ with cosine decay.

\subsection{Joint video fusion and frame interpolation}
\label{sec:joint_results}

\subsubsection{Qualitative evaluation}

Figure~\ref{fig:joint_qualitative} compares:
\textbf{Same rate, fusion then interpolation:} in M3SVD's staircase scene, all baselines brighten the pedestrian's trousers, reducing tonal variation, and distort facial contours. The infrared image connects bright facial and trouser regions. SAFI preserves dark clothing and local structure despite missing target-time frames in both modalities.
\textbf{Same rate, interpolation then fusion:} in the HDO nighttime scene, MAVFusion and UniVF leave black facial artifacts, and VideoFusion produces ghosting around the person. SAFI retains the red top, vehicle colors, water texture, and brightness difference between the water and foreground person.
\textbf{Different rate, interpolation then fusion:} in VTMOT's daytime scene, the visible input has clear human detail. Baselines brighten large areas of black clothing, washing out texture and tonal detail; SAFI preserves black clothing and local details.

\subsubsection{Quantitative evaluation}

Across complete output sequences (Table~\ref{tab:joint_results}), SAFI better preserves source information and perceptually relevant image features than cascaded baselines. These gains hold across all datasets and both frame-rate settings, including both same-rate cascade orders. With different-rate inputs, SAFI improves temporal consistency across all datasets.

\subsubsection{Temporal consistency analysis}
\label{sec:temporal_analysis}

In Fig.~\ref{fig:temporal_consistency}, SAFI's difference maps have red/blue regions closer to white (zero difference) and fewer large dark regions, indicating smaller luminance changes. Motion-compensated errors are lower at most times, with some lower peaks. Local mean luminance shows smaller drops and oscillations than some baselines, remains more consistent between observed and interpolation times, and retains later gradual changes.

\subsection{Fusion performance at observed times}
\label{sec:observed_fusion}

With both modalities observed, SAFI preserves clothing colors and tonal detail obscured by large bright regions in baseline outputs (Fig.~\ref{fig:observed_fusion}a). Human contours and clothing folds are clearer than in VideoFusion's smoother outputs. SAFI leads each spatial metric on all datasets (Fig.~\ref{fig:observed_fusion}b).

\subsection{Computational efficiency}
\label{sec:computational_efficiency}

At $640\times480$ on one NVIDIA RTX 4090, SAFI has the lowest FLOPs and mean runtime per frame (Table~\ref{tab:efficiency}): 45.141~ms (same-rate) and 42.388~ms (different-rate), about $5.4$--$10.8\times$ as fast as each group's fastest cascade.

\subsection{Downstream object tracking}
\label{sec:downstream_tracking}

On VTMOT, SAFI uses inputs with missing frames; baselines fuse complete videos. Official YOLO26 weights and ByteTrack~\cite{zhang2022bytetrack} provide detection and multi-object tracking. We measure HOTA, its detection and association components DetA and AssA~\cite{luiten2021hota}, MOTA~\cite{bernardin2008clear}, and IDF1~\cite{ristani2016identity}. With fewer frames, SAFI leads all five metrics (Table~\ref{tab:tracking}), improving overall tracking, detection, and identity association.

\subsection{Ablation studies}
\label{sec:ablation}

We independently retrain four variants under the full model's settings to test semantic guidance, shared source selection, the output representation, and temporal supervision.
\textbf{A1} removes DINOv3 feature extraction, semantic propagation, and semantic-source fusion, retaining both modalities' image encoding and motion alignment.
\textbf{A2} separates endpoint aggregation weights for interpolation and fusion; interpolation weights remain shared by features and pixel bases within each modality.
\textbf{A3} replaces addition to the visible reference and single-channel thermal enhancement with direct RGB prediction, retaining the multiscale decoder backbone and auxiliary reconstruction supervision.
\textbf{A4} removes the temporal constraints on the enhancement increment and low-frequency fluctuations in the fused luminance error, retaining per-frame supervision.

Across both frame-rate settings, every ablation lowers spatial quality and temporal consistency relative to the full model (Table~\ref{tab:ablation}). A1 has the lowest different-rate SSIM among variants, supporting semantic guidance for joint reconstruction and fusion with different-rate inputs. A2 supports sharing image-source weights between tasks. A3 has the highest MMCI among variants in both settings, supporting bounded thermal enhancement of the visible reference for temporal stability. A4 increases TCPE and MMCI, supporting temporal constraints for inter-frame stability and spatial reconstruction.

\section{Conclusion}
\label{sec:conclusion}
SAFI jointly fuses and interpolates infrared and visible videos in both settings with one set of parameters, using target-time visible semantics and shared image-source weights. Bounded thermal enhancement preserves the visible reference's channel differences; motion-aligned constraints address fluctuations in enhancement increments and low-frequency fused luminance errors while accounting for actual thermal changes. Experiments show better different-rate temporal consistency and same-rate structural preservation with less computation than the evaluated cascades. Despite missing frames, SAFI improves tracking over baselines using complete videos. SAFI requires spatial registration; reference headroom limits thermal enhancement.

\FloatBarrier
{
    \small
    \bibliographystyle{ieeenat_fullname}
    \bibliography{main}
}
\end{document}